\documentclass[conference]{IEEEtran}
\IEEEoverridecommandlockouts
\usepackage{cite}
\usepackage{amsmath,amssymb,amsfonts}
\usepackage{algorithmic}
\usepackage{graphicx}
\usepackage{textcomp}
\usepackage{xcolor}
\usepackage{multirow}
\usepackage{tikz}
\def\BibTeX{{\rm B\kern-.05em{\sc i\kern-.025em b}\kern-.08em
    T\kern-.1667em\lower.7ex\hbox{E}\kern-.125emX}}
\IEEEoverridecommandlockouts

\definecolor{myyellow}{HTML}{FCFD99}

\newcommand\name{VcLLM }
\newcommand\namens{VcLLM}

\newcommand*\circled[1]{\tikz[baseline=(char.base)]{
            \node[shape=circle,draw,inner sep=1pt] (char) {#1};}}

\usepackage[hidelinks]{hyperref}
\usepackage{enumitem}

\begin{document}
\title{\namens: Video Codecs are Secretly Tensor Codecs}

\author{\IEEEauthorblockN{Ceyu Xu\IEEEauthorrefmark{1}\IEEEauthorrefmark{2}}
\and
\IEEEauthorblockN{Yongji Wu\IEEEauthorrefmark{1}\IEEEauthorrefmark{2}}
\and
\IEEEauthorblockN{Xinyu Yang\IEEEauthorrefmark{1}\IEEEauthorrefmark{3}}
\and
\IEEEauthorblockN{Beidi Chen\IEEEauthorrefmark{3}}
\and
\IEEEauthorblockN{Matthew Lentz\IEEEauthorrefmark{2}}
\and
\IEEEauthorblockN{Danyang Zhuo\IEEEauthorrefmark{2}}
\and
\IEEEauthorblockN{Lisa Wu Wills\IEEEauthorrefmark{2}}
\thanks{\IEEEauthorrefmark{1} Ceyu Xu, Yongji Wu, and Xinyu Yang contributed equally to this work.}
\thanks{\IEEEauthorrefmark{2}Department of Computer Science, Duke University, USA; Emails: \{ceyu.xu, yongji.wu769\}@duke.edu, \{mlentz, danyang, lisa\}@cs.duke.edu.}
\thanks{\IEEEauthorrefmark{3}Department of Electrical and Computer Engineering, Carnegie Mellon University, USA; Emails: {xinyuya2, beidic}@andrew.cmu.edu.}
}

\maketitle

\thispagestyle{plain}
\pagestyle{plain}

\begin{abstract}
As the parameter size of large language models (LLMs) continues to expand, the need for a large memory footprint and high communication bandwidth have become significant bottlenecks for the training and inference of LLMs. 
To mitigate these bottlenecks, various tensor compression techniques have been proposed to reduce the data size, thereby alleviating memory requirements and communication pressure.

Our research found that video codecs, despite being originally designed for compressing videos, show excellent efficiency when compressing various types of tensors. 
We demonstrate that video codecs can be versatile and general-purpose tensor codecs while achieving the state-of-the-art compression efficiency in various tasks.
We further make use of the hardware video encoding and decoding module available on GPUs to create a framework capable of both inference and training with video codecs repurposed as tensor codecs. This greatly reduces the requirement for memory capacity and communication bandwidth, enabling training and inference of large models on consumer-grade GPUs. 
\end{abstract}


\section{Introduction}

Recently, Large language models (LLMs) have achieved significant success, showcasing remarkable proficiency in various applications, such as virtual assistants~\cite{gpt4}, chatbots~\cite{gpt4}, and automated customer service platforms~\cite{langchian_custom_service}. 
As their parameter size grows, LLMs develop emergent abilities~\cite{emergent}. 
These abilities enable them to carry out more advanced and complex tasks such as code generation~\cite{starcoder, codellama}, mathematical problem-solving~\cite{mathqa}, and theorem proving~\cite{gpt_theory_prove}, even some they weren't explicitly trained for. 
This growing range of applications has motivated researchers to train increasingly larger models, such as GPT4~\cite{gpt4}, Nemotron-4-340B~\cite{nemotron}, and LLaMA-3-70B~\cite{llama3}, which further leads to a revolution in the development and deployment of AI systems and hardware. 

\begin{figure}
    \centering
    \includegraphics[width=\columnwidth]{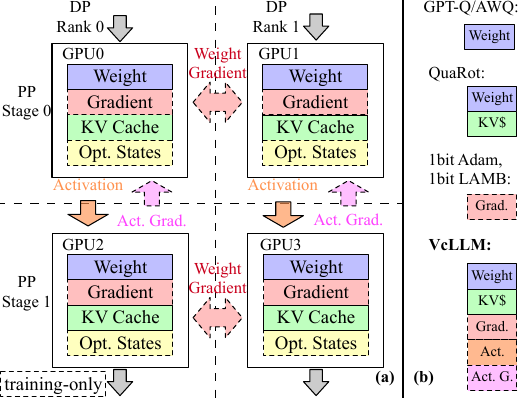}
    \caption{\namens: General-Purpose and Versatile Tensor Compression for LLM Training and Inference.}
    \label{fig:unified_vcllm}
\end{figure}

The training and inference of these large language models with large parameter sizes often strain the underlying computing infrastructure, posing challenges in terms of memory capacity and communication bandwidth. 
For example, inferencing a Nemotron-4-340B~\cite{nemotron} model requires at least 680GB memory, far exceeding the capacity of a single GPU. 
To address this problem, parallelism strategies, such as pipeline parallelism (PP)~\cite{harlap2018pipedream,huang2019gpipe} and data parallelism (DP)~\cite{dean2012large,chen2016revisiting}, have been developed to distribute the models across multiple GPUs and to increase the throughput for inference and training.
Figure~\ref{fig:unified_vcllm} (a) shows an example of a distributed LLM using pipeline parallelism and data parallelism. 
However, inter-GPU communication is required for such a distributed model. 
As shown in Figure~\ref{fig:unified_vcllm} (a), activations need to be transmitted between pipeline stages for distributed inference. During training, the communication pressure is even higher because, in addition to the activations, both the weight gradients and the activation gradients also need to be transmitted across GPUs. 

The requirement for memory capacity and communication bandwidth makes it impractical for most people to train and run their LLMs on commodity-level GPUs (e.g., RTX 3090) with low memory capacity and low communication bandwidth. 
Consequently, users must rely on data centers operated by large companies for LLM training and inference. 
This dependence raises privacy concerns~\cite{sun2024trustllm, yao2024survey} and results in the underutilization of widely available commodity-level GPU resources~\cite{liang2024communication, thorpe2023bamboo}. 
Moreover, even large companies face difficulties further scaling up their data centers due to communication bandwidth and power limitations~\cite{patel_nishball_2024}, hindering their ability to train larger models.
Given these challenges, a critical question arises: \emph{Can we further improve the training and inference efficiency of LLMs in both memory- and communication-constrained environments?}

While scaling up systems and hardware for LLM workloads remains challenging, optimizing data movement and storage becomes an overarching goal, of which compression has become a crucial strategy. 
Compression trades more computation for a reduced amount of data to be maintained, which subsequently translates to a reduced memory footprint and reduced pressure on the communication system.
Figure~\ref{fig:unified_vcllm} (b) lists the tensors required for inference and training of LLMs. 
Traditionally, these tensors are stored, inferred, and trained in half-precision float (FP16) or Brain-float (BF16) \cite{tpuv4} formats, with each value occupying two bytes. 
Research has shown that with compression, model weights can be reduced to 3-4 bits~\cite{gptq, awq} and gradients to 3 bits~\cite{adam1bit, lamb1bit} without significant accuracy degradation. 
However, existing tensor compression techniques also face specific challenges: they are not \textit{general-purpose}~\cite{gptq, awq, adam1bit, lamb1bit} and lack \textit{versatility}. 
As illustrated in Figure~\ref{fig:unified_vcllm} (b), a variety of tensors, such as model weights, gradients, and activations, need to be maintained. 
None of the existing approaches is general-purpose, typically requiring different compression algorithms for each tensor type.
In addition, some approaches~\cite{adam1bit, gptq, awq, smoothquant, lamb1bit} are not versatile due to their reliance on data-aware calibrations and warm-up periods. This dependence complicates deployment and limits robustness when calibration data is unknown or biased.
Moreover, existing algorithms often rely on rounding values to short integers, restricting each compressed value to take an integer number of bits. As a result, achieving a fractional bitrate (defined as the average number of bits in the compressed form per value in the uncompressed form) is impossible, further limiting their versatility.

Our insight is that video codecs (consisting of both an encoder and a decoder), despite being designed for video compression, work well and even achieve state-of-the-art information efficiency for various tensor compression tasks. 
Specifically, we found the distribution of the values representing pixels in videos shares characteristics with tensors in LLMs, allowing video codecs to compress tensors efficiently with only minimal adjustment of codec parameters.
Modern GPUs are equipped with on-chip video encoding and decoding engines (e.g., NvEnc/NvDec on Nvidia GPUs~\cite{videosdk}). 
This allows us to directly leverage these resources for optimal tensor compression throughput. 
We refer to our method of using video codecs for tensor compression as \textbf{V}ideo \textbf{C}oded \textbf{LLM} (\namens).
\name is general-purpose: it offers a unified compression method that effectively compresses various types of tensors while achieving state-of-the-art information efficiency across all tasks.
\name is also versatile: being data-independent, \name requires no data-aware calibration or warm-up.
In addition, \name works at fractional bitrates (e.g. 2.3 bits per value), not limited to integer bitrates.
Thanks to these properties, \name allows for extreme LLM compression by simultaneously compressing multiple types of tensors, while providing the capability of fine-grained fractional bitrate tuning for ultimate information efficiency.
The reduced memory footprint and communication bandwidth requirement enable the training and inference of large models on commodity-level GPUs with limited resources. 
We demonstrate that \name is the first method capable of conducting inference for the LLaMa-3-70B~\cite{llama3} model with a sequence length of 128k on 4 $\times$ 8GB devices, using 3.5 bits per value for communication and 2.9 bits per value for weight and KV cache compression. 
In contrast, previous methods only compress weight and KV cache to 4 bits at most and do not consider activation compression for communication.

In this paper, we will first provide empirical evidence demonstrating the effectiveness of video codecs in tensor compression. 
Our experiments then showcase how \name can compress the weights, KV cache, and activations of LLMs, thereby reducing the memory footprint and communication cost during inference.
To highlight the general-purpose capability of \namens, we further show its efficacy in compressing gradients to reduce communication size during distributed training under pipeline and data parallelism.
Finally, we will discuss the insights \name offers for future GPU and accelerator hardware design, exploring the potential for developing custom hardware codecs specifically tailored for tensors.

We make the following contributions:
\begin{enumerate}[leftmargin=*]
    \item We demonstrate that video codecs such as H.264 and H.265 are highly effective for compressing various types of tensors, including weights, activations, and gradients.
    \item We present empirical evidence demonstrating the effectiveness of video codecs for tensor compression, providing valuable insights to guide the development of future tensor compression algorithms.
    \item We develop \name that leverages the hardware video codecs in modern GPUs to compress tensors during LLM inference and training. It achieves compression ratios of up to 3-20$\times$ while delivering superior throughput compared to state-of-the-art compression methods.
    \item We further show that video codecs can be  augmented into more efficient \textit{tensor codecs} specialized for tensor compression, reducing die area and power consumption.
    We propose the integration of tensor codecs into future GPU and accelerator System-on-Chip (SoC) designs.
\end{enumerate}

\section{Background}
\subsection{Model Compression}

Quantization techniques have been widely applied in model compression~\cite{han2015deep, Jacob_2018_CVPR, banner2018scalable, nagel2019data}. Most current quantization methods are based on vanilla round-to-nearest quantization (RTN). Given a tensor $T$, the RTN quantization function is \mbox{defined as:}
 
 \begin{equation}
Q(\mathbf{w}) = \Delta \cdot \text{Round}\left(\frac{\mathbf{w}}{\Delta}\right), \quad \Delta = \frac{\max(|\mathbf{w}|)}{2^{N-1}},
\end{equation}
where \(N\) is the number of quantization bits, and \(\Delta\) is the quantization scaler determined by the absolute maximum value. Additionally, recent work has also explored non-uniform quantization techniques such as K-means clustering~\cite{kim2023squeezellm}, vector quantization~\cite{tseng2024quip}, and NormalFloat quantization~\cite{qlora}. While our work primarily focuses on dense compression methods, several other studies have explored sparse model compression for both training and inference~\cite{zhang2024h2o, song2023optimus, sun2023simple}. These methods are orthogonal to our methods. We leave the discussion and comparison of these work for future research.

\textbf{Weight Compression: }
In LLM inference, the size of model weights can be a bottleneck.
Modern LLMs can scale up to 340 billion parameters~\cite{nemotron}, necessitating 680 GB of GPU memory. This would require distributing the model across up to 9 Nvidia H100-80GB GPUs, creating substantial communication overheads. To address this issue, LLMs are usually compressed with quantization-aware training (QAT) methods~\cite{liu2023llm, qlora, xu2024onebit} or post-training quantization (PTQ) techniques~\cite{gptq, awq, quip, tseng2024quip}. 
QAT approaches require additional training, while PTQ algorithms calibrate the rounding of weights into low-bit integers by running the model through a calibration dataset. Through this calibration process, weights can be compressed from 16 bits to 3-4 bits, while still maintaining acceptable model accuracy.

\textbf{Activation Compression: }
Simply compressing the model weights is not sufficient. 
During inference, especially in scenarios involving long-context lengths and large batch sizes, the size of the activations (including key-value (KV) cache) also becomes a bottleneck. 
Dual-side quantization and compression techniques, such as SmoothQuant~\cite{smoothquant} and QServe~\cite{qserve}, has been developed to compress both model weights and activations. Activation compression presents a greater challenge due to the significant outliers that exist in the activation's distribution~\cite{smoothquant}. As a result, current activation compression algorithms typically achieve 8-bit compression without accuracy loss~\cite{qserve}. Some methods reach 4-bit compression~\cite{quarot, spinquant}, but often with degraded accuracy. These approaches primarily focus on compressing the activations before matrix multiplication in linear layers, thereby accelerating the GEMM Kernel on modern GPUs.
However, none of them explore compressing activations transmitted between machines to reduce the communication cost, which is much more time-consuming in distributed settings. In our work, we consider compressing the KV cache during inference, as well as the activations between different pipeline parallelism stages in both training and inference. 

\textbf{Gradient Compression: }
Model compression in training becomes more complex during backward propagation stages, 
where maintaining gradients is a major bottleneck for memory and communication. In distributed training, gradients need to be exchanged across machines, often causing communication bottlenecks.
1-bit Adam~\cite{adam1bit} and 1-bit LAMB~\cite{lamb1bit} compress the weight gradient to an average of 3-4 bits using a two-stage approach. 
In the warm-up stage, 16-bit floating point values
are transmitted without compression
, as the model hasn't converged to a point where the weights can be easily compressed yet.
Once the model convergence becomes stable, these algorithms enter a variance-freeze stage, where they can compress the weight gradients to 1 bit per value.
Notably, these methods only support weight gradient compression while not supporting activation gradient compression, limiting their use-case to data parallelism but not pipeline parallelism as pipeline parallelism
requires communication of activation gradients.
In our work, we apply our method to compress both weight gradients in data parallelism and activation gradients in pipeline parallelism.

\textbf{Problems of Existing Approaches: }
We have identified two main issues of existing approaches.
First, existing tensor compression algorithms are not general-purpose, with each algorithm only capable of compressing one or two types of tensors. 
This makes it complex to create a "compress-everything" system, and the quality of the results when using these algorithms in conjunction is unverified. 
Second, all these algorithms require specific data-dependent calibrations and parameter tuning, making the system design more complex and raising concerns about their robustness across different models and varying data distributions.

\subsection{AVC and HEVC Video Codecs}
\begin{figure*}
    \centering
    \small
    \includegraphics[width=\textwidth]{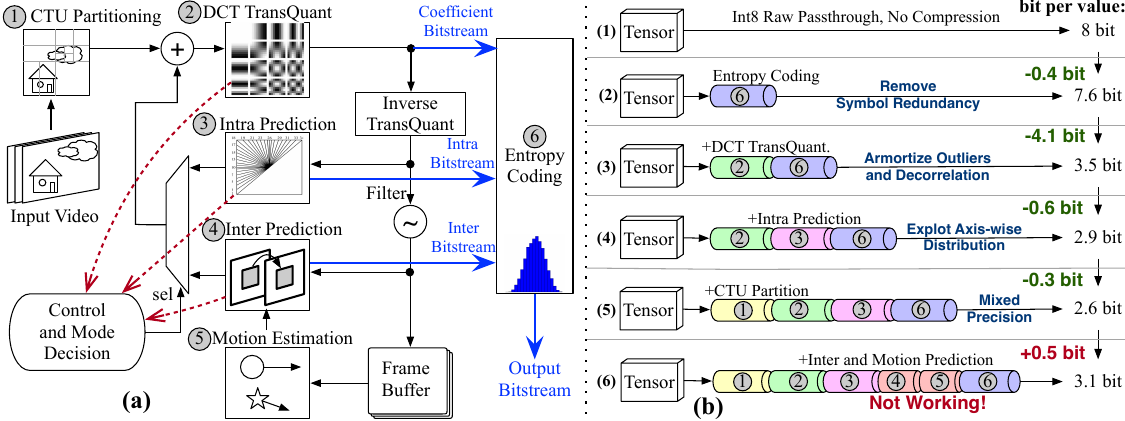}
    \caption{Why does the Video Codec Work for LLM? (a) illustrates the pipeline of the H.265 video encoder. In (b), we incrementally activate the stages in the H.265 video encoding pipeline to demonstrate how each step contributes to the compression process. We constrain the quality of the compression/decompression process to have a maximum \mbox{mean square error of 0.01.}}
    \label{fig:llm_video_codec}
\end{figure*}

Video codecs are essential techniques that empower everyone's daily media consumption.
Modern encoding techniques such as Advanced Video Coding~\cite{h264_standard} (AVC or H.264) and High-Efficiency Video Coding~\cite{h265_standard} (HEVC or H.265) enable efficient video streaming over limited network bandwidth by compressing raw video footages up to a ratio of 1000:1 with unnoticeable quality loss of the videos.
A video codec consists of an encoder that compresses raw videos to compressed bitstreams and a decoder that reconstructs every pixel from the bitstreams.
The H.264 and H.265 standards establish a set of fixed rules for the decoding process while allowing flexibility in implementing the encoding process.
The encoding pipeline is a compounding of several unique compression blocks. 
Figure~\ref{fig:llm_video_codec} (a) shows a typical H.265 encoding pipeline implementation.
The encoding process begins with raw video frames. 
First, a process called the \circled{1}CTU (code tree unit) partitioning divides the video into a Quad-Tree of Coding Units (CUs)~\cite{hevc_overview}. 
Then, predictability between \circled{3}intra-frame, and \circled{4}\circled{5}inter-frame pixels in each CU is utilized, so if some pixels can be well-predicted, we no longer need to store them. 
The residual between the actual pixels and the prediction is measured, transformed, quantized, and stored as coefficients. 
This step is called the \circled{2}DCT transform. 
Finally, all the predictive states, coefficients, and meta-data are input into an \circled{6}entropy coder (e.g., CABAC~\cite{cabac}) to exploit system-level symbol redundancies.

\section{Video Codecs are secretly Tensor Codecs}\label{sec:codec_for_tensor}
\subsection{Why do Video Codecs Work for Tensor?}
Video codecs achieve high-quality and efficient compression by leveraging prediction.
The idea is that the majority of pixels can be predicted, leaving only sparse and small residuals (differences between the actual frame and the prediction) to be encoded.
In addition to the prediction, steps such as Discrete-Cosine Transform (DCT), quantization, and entropy coding exploit other types of redundancies in the video that are either imperceptible to the human eye or can be masked due to the non-uniform distribution of symbols.

Our work demonstrates that some stages in the video coding pipeline are also effective for compressing tensors. 
To analyze why video codecs work and how each stage in the pipeline contributes to the compression of large language model tensors, we set up an experiment where we enabled the stages in the encoding pipeline step-by-step, as shown in Figure~\ref{fig:llm_video_codec} (b) from (1) to (6).
The video codecs take a 4D input, with dimensions representing time, color channel, width, and height. 
In our experiment, we used the weight tensor of the Key-Projection linear layer in the LLaMA-2-7B~\cite{llama2} network as an example. 
We constructed a 4D video tensor from the 2D weights of the Key-Projection linear layer, using the layer index as the temporal channel and only the Luma channel for gray-scale encoding, with Chroma channels padded with zeros.
Video codecs like H.264 and H.265 allow users to set the bitrate target explicitly. 
We constrained the maximum distortion to a mean square error (MSE) of less than 0.01.
Detailed analysis of how the bitrate of the codec and the distortion of the weight will affect the LLM's accuracy will be shown in Section~\ref{sec:weight_only}. \mbox{Here, we use $\text{MSE}<0.01$ as an example.}
We sweep the bitrate from low to high for each codec pipeline setting until we find a bitrate that achieves this quality constraint. 
We demonstrated that incrementally activating stages in this pipeline reduced the average bits per value from 8 bits to 2.6 bits for achieving a quality of $\text{MSE}<0.01$.

\textbf{Entropy Coding: }
Entropy coding is a lossless compression technique used in various compression algorithms. It assigns shorter codes to more frequent symbols and longer codes to less frequent symbols. 
In the context of video codecs, entropy coding can exploit redundancies in the distribution of symbols, reducing data size without introducing additional distortion due to its lossless nature.
The effectiveness of entropy coding in compressing videos can be generalized to compressing tensors. 
As prior works have shown, weights, activations, and gradients in LLM training and inference all conform to a normal or bell-shaped distribution~\cite{fp4, awq, qlora}. 
The non-uniformity in symbol distribution allows entropy coding to achieve an average reduction of 0.4 bits per value for the weight tensor, as illustrated in Figure~\ref{fig:llm_video_codec} (b) (2).

\begin{figure}
    \centering
    \small\includegraphics[width=\columnwidth]{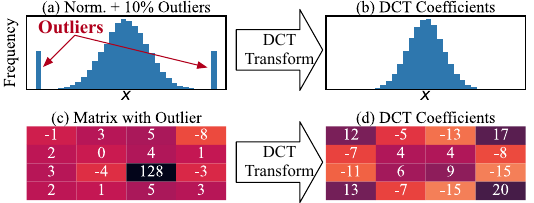}
    \caption{Transform coding mitigates encoding outliers by mapping them to all values within the block. The transition from (a) to (b) demonstrates how DCT removes outliers from a normal distribution matrix containing outliers. 
    (c) to (d) shows a concrete example of how an outlier with a value of 128 is "smoothed" into other values within the block.}
    \label{fig:transform_coding}
\end{figure}

\textbf{Transform Coding: }
Transform coding is a vital technique used in video and image codecs. 
It automatically de-correlates pixels in the frame and removes high-frequency information that is less perceptible to human eyes. 
One popular technique in transform coding is the DCT transform~\cite{dct}, which is utilized in both H.265 and H.264. 
The DCT transform is based on the fact that for an orthogonal basis matrix $\mathbf{B}$ and input $\mathbf{X}$, the encoding process can be represented as $\mathbf{Y} = \mathbf{X}\mathbf{B}$, and the encoded $\mathbf{Y}$ can be decoded by $\mathbf{X} = \mathbf{Y}\mathbf{B^{-1}}$.

In tensor compression, however, if we view tensors as images, the emphasis on low-frequency signals through DCT transform may visually preserve similarity, but this does not translate to improved compression quality in terms of the accuracy of model-specific tasks (e.g., the quality of the generated sentences). 
Instead, transform coding is effective in tensor compression for a different reason --- it mitigates the encoding difficulties caused by outliers in tensors~\cite{quarot, slicegpt}.
Outliers that are far away from the center distribution, sometimes even degrees of magnitude different from centered values, put conventional quantization and compression techniques in a dilemma that could either encode the outliers but leave the center distribution's encoding in low resolution or clip the outliers to better adapt to the range of the center distribution, but not both. 
Prior works~\cite{smoothquant, llmint8} showed that suppressing the outliers or decreasing the center quantization granularity both decreases the accuracy of LLMs. 

However, transform coding solves this dilemma.
In Figure~\ref{fig:transform_coding}, we show the effect of the DCT transform: 
(a) exemplifies a common tensor distribution where the central distribution is near-normal, but outliers exist at both tails.
The DCT transform solves the challenge of encoding outliers; its output, as seen in (b), no longer contains outliers. 
A more concrete example is shown in the process from (c) to (d), where we can see that the value of 128 is an outlier in (c). 
The DCT transform addresses this by amortizing the difficulty of encoding the outlier value 128 to other values within the same block. 
This results in a matrix (d) containing no outliers and is much easier to encode in binary space.

\begin{figure}
    \centering
    \small
    \includegraphics[width=\columnwidth]{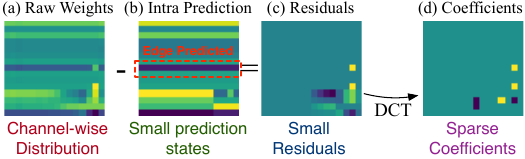}
    \caption{An example of a block of LLaMA-2-7B~\cite{llama2} weights going through the H.265 pipeline. The intra-prediction step generates a rough prediction of the entire block, making the residuals easy to code with the DCT transform.}
    \label{fig:pipeline_example}
\end{figure}

\textbf{Intra-Frame Prediction: }
Intra-frame prediction is another crucial component in modern video codecs.
It is based on the simple fact that objects in the frames can be predicted or approximated through a few classes of patterns. 
For example, smooth areas of the frame can often be approximated by predicting the pixel values from neighboring pixels using a planar or DC (direct current) prediction mode. 
Edge areas, which are common in real-world images, can be predicted using directional modes that capture the orientation of edges. 
Although it is usually impossible to predict the pixels in a block with very high accuracy, residual encoding can be used to improve the quality further. 
As long as the prediction is close to the original block, the residuals will be small in size and much easier to encode compared \mbox{to the raw values.}

To our surprise, the intra-frame prediction works well for compressing tensors. 
We present an example in Figure~\ref{fig:pipeline_example}. 
In (a), a block of a weight tensor is depicted as an image.
We then performed H.265 encoding on this image as a frame, extracting the prediction from the intra-frame predictor of H.265\footnote{The prediction states are extracted using the HEVC Test Model (HM)~\cite{hm}.}. 
The predicted image is shown in (b), and the residual, which is the difference between the original block and the prediction, is shown in (c).
We made three observations for applying intra-frame prediction for weight images.
First, the original weight, when viewed as images, contains edges and planar blocks that are similar to real-world images due to the channel-wise distribution property, as shown in prior works~\cite{awq, smoothquant, quarot}. 
The channel-wise distribution property means each value's distribution aligns with the corresponding channel, causing values close to each other to appear within the same channel, which visually looks like the edges of objects.
Second, the intra-frame prediction mechanism can detect the channel-wise distributions and efficiently encode them using small-sized prediction states. 
Third, the residuals after the intra-frame prediction are much smaller in size compared to the original weight distribution and require much fewer states to properly encode.
This, when used in conjunction with Transform coding and quantization as shown in Figure~\ref{fig:pipeline_example} (d), results in sparse and small coefficients that are very easy and efficient to encode using \mbox{only a few bits.}

\textbf{Inter-Frame Motion Prediction Does not Work}
\label{sec:algo_inter_frame}
Although the Inter-Frame prediction, including the motion prediction, achieves great efficiency in compressing videos, based on our experiments, it does not work for compressing tensors. 
As shown in Figure~\ref{fig:llm_video_codec} (b) (5) $\rightarrow$ (6), enabling the inter-frame prediction stage does not help reduce the number of bits per value but rather increases it. 
This observation suggests there is little inter-frame pixel correlation and little inter-layer correlation of weights in LLMs.
Consequently, for all subsequent experiments in this paper involving the use of video codecs to compress tensors, we configure the codec parameters to disable the inter-frame prediction stage.

\subsection{\name Implementation}
\label{sec:impl}

We implement \name on top of the PyTorch~\cite{pytorch} framework.
Modern GPUs contain specialized hardware codecs designed to encode and decode videos. 
\name utilizes NVENC and NVDEC, which are the hardware video encoders and decoders present on NVIDIA GPUs~\cite{videosdk}. 
As NVENC and NVDEC have a maximum limit on the height and width of a frame, we first partition each input tensor into multiple chunks, each corresponding to a frame.
Since video codecs take 8-bit integers as input, the FP16 values in the tensor need to be first rounded to 8 bits using RTN quantization before feeding to HEVC codec, with only Luma channel of the codec is used. 
As mentioned in Section~\ref{sec:algo_inter_frame}, inter-frame compression is ineffective; therefore, \name enforces an intra-frame-only encoding by setting codec parameters.

\section{Memory- and Communication-Efficient Inference Using \name}
\label{sec:weight_only}
Built upon the \name implementation in \autoref{sec:codec_for_tensor}, we begin to improve the memory and communication efficiency in LLM inference, where our goal is to run a LLaMA-3-70B~\cite{llama3} model with 128k context length on 4 edge devices with only 8GB memory. This challenging objective requires a general-purpose compression strategy including three critical steps:

\textbf{Weight Compression.} In \autoref{weight_compression}, we show that \name can reduce the memory footprint of model weight by 5.5$\times$ while maintaining accuracy. Notably, this is accomplished by compressing the weights from 16 bits to 2.9 bits without any calibration or training, making it entirely data-independent. Our weight compression enables us to run a LLaMA-3-70B model with only about 25GB of memory.

\textbf{KV Cache Compression.} In \autoref{kv_compression}, we employ \name to compress the KV cache to 2.9 bits without degrading accuracy. This reduces the cache size for a 128k length context from 40GB to 7.2GB for the LLaMA-3-70B model.

\textbf{Communication Compression.}  In \autoref{kv_compression}, we distribute the model across 4 devices using pipeline parallelism and compress the activations between different stages using \namens. By reducing the bit-width to 3.5 bits, our method can speed up the communication by 4.5 times.

\subsection{Weight Compression}
\label{weight_compression}

In this subsection, we show our \name as the first \textit{data-independent} method for low-bit (i.e., $\leq 3$ bits) LLM weight compression. \name is versatile and accurate, and it is \textit{outlier-free, calibration-free, and training-free}. Compared to existing quantization techniques, these features significantly improve efficiency and robustness in compressing large models. 

The need for such a compression method arises from the challenges of deploying LLMs on memory-constrained devices, particularly when these models require further training and adaptation for specialized tasks by end-users.  
 For low-bit weight quantization methods, quantization-aware training (QAT)~\cite{liu2023llm, xu2024onebit} is computationally expensive due to the high training cost. Conversely, post-training quantization  (PTQ)~\cite{awq, gptq} is more efficient but heavily depends on the calibration set, which can limit its generalization ability across diverse models and tasks~\cite{outliers}. As a result, neither QAT nor PTQ can be considered truly zero-shot methods since they involve a ``fine-tuning" step, raising concerns about their efficiency and robustness in real-world applications.

\subsubsection{Two-stage Compression Strategy}

As detailed in \autoref{sec:impl}, NVENC only supports 8-bit integers as input, while the weights are stored in FP16 or BF16 precision.
To address this discrepancy while minimizing compression errors, we develop a two-stage strategy for the weight compression. 

\textbf{RTN Quantization with Incoherent Processing.} Our first stage quantizes the model weights to 8-bit integers.To maintain data independence and generalizability, we use RTN quantization, a vanilla rounding quantization method without calibration or training.  To further even out outliers, we apply the incoherence processing described in QuIP~\cite{quip}. This involves applying rotation matrices to both sides of each weight matrix before quantization, which helps to ``spread out" the outliers in the weights and makes quantization easier. In practice, we 
use the method proposed in QuaRot~\cite{quarot} and choose the rotation matrices to be the product of random diagonal matrices of $\pm1$ and a Hadamard matrix. These randomized Hadamard matrices are a specific class of orthogonal matrices that perform well in filtering outliers~\cite{quarot, spinquant}. 

To reduce the inference overhead,  
we also merge the rotation matrices into the original weights following these work. This is achieved by using the computational invariance theorem~\cite{slicegpt}. Taking the LLaMA model as an example, each attention block and feed-forward network block in the architecture includes linear operations on both its input and output, represented by matrices $\mathbf{W}_{\text{in}}$ and $\mathbf{W}_{\text{out}}$ respectively. To remove outliers in $\mathbf{W}_{\text{out}}$, we multiply it on the right by an orthogonal matrix $\mathbf{P}$, resulting in a new weight matrix $\mathbf{W}_{\text{out}}\mathbf{P}$. Since $\mathbf{W}_{\text{out}}$ and $\mathbf{W}_{\text{in}}$ are connected, we can counteract this effect by multiplying $\mathbf{W}_{\text{in}}$ with $\mathbf{P}^{\top}$ on the left. Given that $\mathbf{P}\mathbf{P}^{\top} = \mathbf{I}$, we maintain the original transformation as $\mathbf{W}_{\text{out}}\mathbf{P}\mathbf{P}^{\top}\mathbf{W}_{\text{in}} = \mathbf{W}_{\text{out}}\mathbf{W}_{\text{in}}$, thereby filtering outliers in weights without introducing new parameters in the model.

\textbf{Variable Bit-Width Compression with Video Codecs.} In the second stage, we use \name{} to further reduce the output from stage one to a low bit-width (i.e., 2-3 bits on average). Our method not only offers high-quality compression but also features fractional 
and variable bit-width, enhancing its versatility and accuracy. Instead of quantizing the model to a fixed integer bit, we can perform a fine-grained search to maintain different compression ratios for different weight matrices.
Additionally, each weight matrix can be adaptively compressed to mixed precision using CTU Partition, and stored in a bitstream format, thus avoiding the hardware inefficiencies of mixed-precision implementation.

\subsubsection{Experiments}
\label{sec:weight_only_exp}

We conducted experiments on the LLaMA-2-7B~\cite{llama2} and LLaMA-3-70B~\cite{llama3} models. Compared to SOTA quantization methods requiring calibration, \name achieves on-par accuracy with higher compression ratios.

\begin{figure}[!h]
    \centering
    \small    \includegraphics[width=\columnwidth]{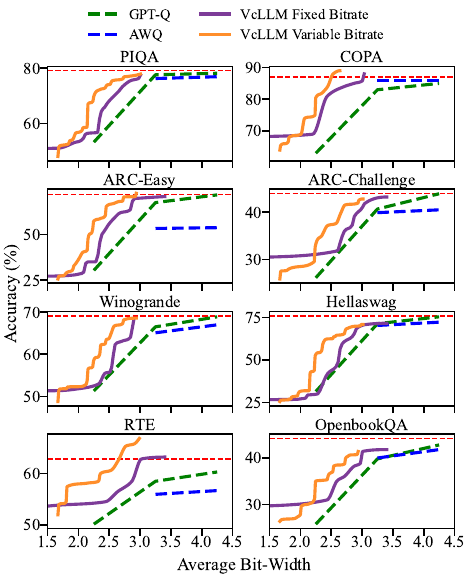}
    \caption{The trade-off between accuracy and average bid-width of different methods for compressing the LLaMA-2-7B model~\cite{llama2} on eight commonsense reasoning tasks.}
    \label{fig:weight_only_eval}
\end{figure}

\begin{table}[!h]
    \centering
    \caption{ Accuracy and average bits of LLaMA-3-70B~\cite{llama3} after compression using different algorithms. "128G" denotes quantizing each group of 128 values separately. }
    \resizebox{\linewidth}{!}{
    \begin{tabular}{cccccc}
    \hline
    \hline
    \# Avg. Bits  & Algorithm  & PIQA & WinoGrande & HellaSwag\\
    \hline
16 & -- & 82.4 & 80.6 & 66.4\\
    \hline
\multirow{2}{*}{3.25} & GPTQ-128G & 80.6 & 77.1 & 63.5\\
 & AWQ-128G & 81.4 &   \textbf{78.6} & 63.5\\
    \hline
\multirow{2}{*}{3.00} & GPTQ & 79.5 & 66.1 & 62.8 \\
 & AWQ & 80.1 & 67.6 & 62.5 \\
    \hline
2.88 & \name(\textbf{Ours}) & \textbf{81.5} & 77.5 & \textbf{63.7} \\
    \hline
    \hline
\end{tabular}}
\label{scale_model}
\end{table}

\textbf{Experimental Setup.} Our evaluation of the proposed \name was carried out on eight zero-shot commonsense reasoning tasks using the LM Evaluation Harness~\cite{eval-harness}. These tasks include PIQA~\cite{piqa}, COPA~\cite{copa}, ARC-easy and ARC-challenge~\cite{clark2018think}, WinoGrande~\cite{sakaguchi2021winogrande}, HellaSwag~\cite{zellers2019hellaswag}, RTE~\cite{glue}, and OpenbookQA~\cite{mihaylov2018can}. For baselines, we compared \name with two state-of-the-art quantization methods: GPTQ~\cite{gptq} and AWQ~\cite{awq}. As these baselines were calibrated using a few samples from WikiText-2~\cite{merity2016pointer}, we excluded the measurement on the calibration dataset from our experiments. 

\textbf{LLaMA-2-7B.} In \autoref{fig:weight_only_eval}, we compare \name and its fixed bitrate variant against other baselines. Our method significantly outperforms all baselines, maintaining full precision accuracy with approximately 3 bits. In contrast, GPTQ and AWQ achieve similar accuracy with around 4.25 bits. Additionally, these baselines struggle to keep accuracy under 3 bits, while \name generalize well to 2.5 bits. Another observation is that \name outperforms its fixed bitrate variant by a large margin in the extremely low bit-width regime (i.e., $< 3$ bits). This validates that different components in LLMs vary in their compression difficulty, and setting different bit widths can further push the limitations of compression.

\textbf{LLaMA-3-70B.} To verify the scalability of our method, we present the results of the compressed LLaMA-3-70B model on three datasets in \autoref{scale_model}. Our method achieves similar accuracy to GPTQ-128G and AWQ-128G with 0.37 fewer bits, and outperforms the 3-bit baselines without groupwise quantization with a large margin. Here, \name benefits from its fine-grained bit-width feature, reducing the bit-width to as low as 2.88 bits. In contrast, prior methods are limited to integer bit-widths with separate groups for quantization, making them less flexible. However, comparing to the results of LLaMA-2-7B, the gap between \name and the baselines narrows as model size and data volume increase, highlighting the importance of calibration in large-scale settings. Since our approach is orthogonal to these algorithms, combining them may yield improved compression results in the future.

\subsection{KV Cache and Communication Compression}
\label{kv_compression}

While our weight compression results show the ability of serving a 70B model on a single commodity-level GPU, it still suffers from two limitations described below.

\textbf{Long-context scenarios:} The large memory requirements of KV cache poses challenges~\cite{zhang2024h2o, sun2024triforce, dong2024get} for long-context LLMs. For example, storing a 128k KV cache using FP16 requires 40 GB of GPU memory for the LLaMA-3-70B model, which is larger than the compressed model itself.

\textbf{On-device Inference:} It is infeasible to run inference for  a 70B model on an edge device with only 8 GB memory.

We address these challenges by applying \name to KV cache and communication compression, enabling distributed inference for LLMs in long-context, on-device scenarios.

\subsection{Experiments}
\label{sec:inference-experiment}

\begin{figure}
    \centering
    \includegraphics[width=\columnwidth]{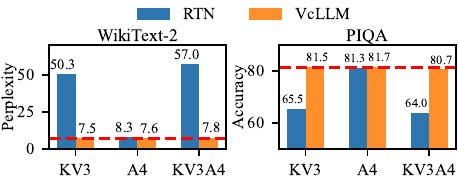}
    \caption{The comparison between RTN quantization and \name for  compressing KV cache and activations of LLaMA-3-70B~\cite{llama3}. "KV3" means compressing KV cache to 3 bits (2.9 bits for \namens), while "A4" compresses activations between pipline stages to 4 bits (3.5 bits for \namens). Lower perplexity and higher accuracy indicate better compression quality.}
    \label{fig:distributed_inference}
\end{figure}

Here, we detailed our final results that reduce the memory footprint by 5.5$\times$ and communication volumes by 4.5$\times$ for the LLaMA-3-70B model using \namens, which only lead to a minor accuracy drop ($< 2\%$) in the zero-shot reasoning task. 

Building on our compressed LLaMA-3-70B model in \autoref{sec:weight_only_exp}, we further compress the KV cache to 2.9 bits and the activations between different pipeline parallelism stages to 3.5 bits. Similarly, we also first compress them to INT8 format using channel-wise RTN quantization.  For baseline comparisons, since these values are dynamically determined at runtime and cannot be preprocessed in advance, we employ RTN quantization to directly reduce the KV cache and activations to 3 bits and 4 bits, respectively, using asymmetric min-max dynamic quantization. Our evaluation includes measuring  the perplexity score on the WikiText-2 test set and the zero-shot accuracy on the PIQA dataset. The perplexity score measure an LLM's fluency and coherence by quantifying its uncertainty in token prediction.

As shown in \autoref{fig:distributed_inference}, our method results in only a 7\% increase (from 7.28 to 7.77) in perplexity score on WikiText-2  and a 1\% drop (from 81.5\% to 80.7\%) in accuracy on PIQA while compressing the KV by 5.5 times and the activations by 4.5 times. Consequently, when the model is distributed across four devices using pipeline parallelism, only about 6.3 GB of memory is required for the compressed model and 1.8 GB for the stored KV cache. This amounts to approximately 8 GB of memory per device. Compared to RTN quantization, we observe that directly quantizing the KV cache to 3 bits leads to a significant accuracy drop, nearly destroying the original model's ability. For activation-only compression, our method achieves only a 5\% increase in perplexity score while RTN quantization results in a 13\% increase.

\section{Communication-Efficient Distributed Training Using \name}\label{sec:dist_training}
In this section, we shift our focus from inference to a more challenging setting: training. The rapid growth of LLMs in both size and data volume has pushed training beyond the memory and computation capabilities of a single node. To address this, we distribute model parameters, activations, gradients, and optimization states across multiple nodes during training using various parallel strategies. However, as systems scale, communication costs increase significantly, accounting for 30\% to 95\% of the total training~\cite{adam1bit,wang2023topoopt,wang2022overlap}.

Consequently, compressing communication is of critical importance, especially on commodity hardware with limited bandwidth.
We demonstrate that \name effectively compresses various tensor types in two parallel-training scenarios, showcasing its versatility and broad applicability.

\textbf{Pipeline Parallelism.} In \autoref{sec:pipeline_parallel}, we show that \name can compress both activations and their gradients between different 
pipeline stages. While prior work has explored sparse compression techniques~\cite{song2023optimus}, our method represents the first \textit{dense compression} solution in this scenario.  

\textbf{Data Parallelism.} In \autoref{sec:data_parallel}, we leverage \name to compress the gradients of weights aggregated across GPUs. Unlike 1-bit Adam \cite{adam1bit} and 1-bit LAMB \cite{lamb1bit}, \name avoids the need for full-precision warm-up and optimizer modifications, leading to more \textit{efficient and stable} distributed training.

For experiments, we build a prototype training system based on DeepSpeed~\cite{rasley2020deepspeed}~(v0.14) using our \name implementation in \autoref{sec:impl}.  We also implement a collective primitive to all-reduce compressed gradients in data-parallel training. All evaluations are performed \mbox{on four RTX 3090 GPUs.}

\subsection{Pipeline-parallel Training}
\label{sec:pipeline_parallel}
 \begin{figure*}
\centering
\includegraphics[width=\linewidth]{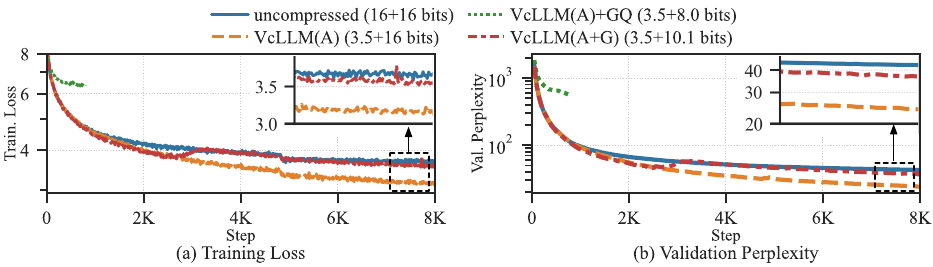}
\caption{Training loss and validation perplexity of Pythia 1.4B using pipeline parallelism. } 
\label{fig:train_pp}
\end{figure*}

\begin{figure*}
\centering
\includegraphics[width=\linewidth]{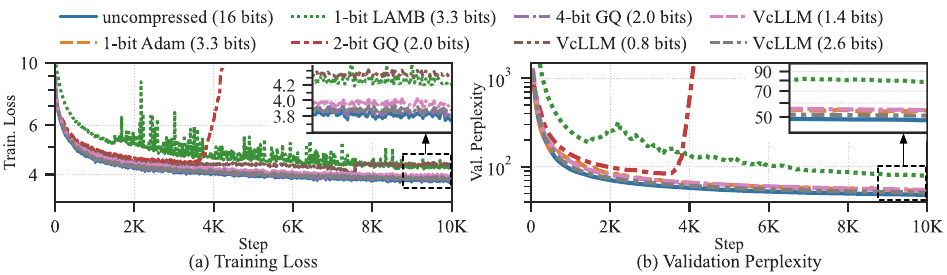}
\caption{Training loss and validation perplexity of Pythia 160M using data parallelism.}
\label{fig:train_dp}
\end{figure*}

Following \autoref{kv_compression}, we further demonstrated the effectiveness of \name in pipeline-parallel training, a predominant method for training large models that exceed single GPU memory capacity. Our approach achieved significant compression ratios: 78\% for activations and 37\% for their gradients when communicating between pipeline stages.

\textbf{Experimental Setup.} We trained a 1.4B Pythia~\cite{biderman2023pythia} model using 4-stage pipeline parallelism across 4 GPUs, with compressed communication between distinct pipeline stages. Our training configuration used a sequence length of 2048, a micro-batch size of 4, and 8 gradient accumulation steps. The model was trained using FP16 precision with the same optimizer settings as in the Pythia repository. We utilized a 5M-sample subset of the Pile dataset~\cite{gao2020pile}, reserving 5000 samples for validation and the rest for training.

\textbf{Activation Compression.} In \autoref{fig:train_pp}, we first verify the transfer of effective activation compression from inference (\autoref{sec:inference-experiment}) to training using \namens(A), where we compress the activations to 3.5 bits. Compared to uncompressed training, \namens's activation compression is surprisingly beneficial. It not only reduces communication volume by 78\% (from 16 bits to 3.5 bits) but also leads to faster convergence. This is evidenced by lower training loss and validation perplexity after 8K training steps (e.g., a validation perplexity of 24.1 compared to 42.7 for uncompressed training). We hypothesize that this improvement stems from \name acting as a denoising operation, filtering out noisy components in the activations and clipping the outliers in the corresponding weights' gradients during backpropagation~\cite{lin2017deep, zhang2019gradient}.

\textbf{Gradient Compression.} To further enhance communication efficiency, we compress the gradients of activations in \namens(A) + GQ and \namens(A+G), as illustrated in \autoref{fig:train_pp}. However, our experiments with \namens(A) + GQ reveal that gradients are more challenging to compress. Even directly applying a group-wise 8-bit RTN quantization to gradients proves ineffective, as the loss deviates from uncompressed training after only a few hundred steps. To address this issue, we introduce a residual compensation method for gradient compression. First, we compress the gradient $G$ to approximately 3.5 bits, denoted as $\operatorname{Comp}(G)$. Next, We further compress the residual $G-\operatorname{Comp}(G)$ using a two-stage strategy with different compression ratios: a) For the first 2500 steps, we use \name to compress the residual to 3.5 bits, achieving a loss curve similar to activation-only compression, and b) After 2500 steps, we switch to 8-bit RTN quantization for the residual. 
This two-stage approach is necessary because the training loss fails to continue decreasing after 2500 steps when using a 3.5-bit residual. This stagnation occurs because the range variance in gradients progressively increases from 1 to 3 orders of magnitude
as training progresses, with some dimensions contributing significantly more to the loss.
By employing this strategy, we achieve an average of 10.1 bits for the compressed gradient, calculated as $((3.5+3.5)*2500 + (3.5+8)*5500)/8000$. Still, an overall compression rate of 37\% in gradient is achieved (from 16 bits to 10.1 bits) and the final validation perplexity is 36.7, which is lower than that of full-precision training.

\subsection{Data-parallel training}
\label{sec:data_parallel}
Next, we show \name's ability to compress weight gradients communicated between GPUs to 1.4-2.6 bits from the starting of data-parallel training without modifying optimizers.

\textbf{Experimental Setup.} We trained the 160M Pythia model with a per-GPU batch size of 8. Following our setup in pipeline parallelism, we adopt the same dataset and use FP16 with optimizer settings provided in the Pythia repository.

\textbf{Results and Analysis.} In \autoref{fig:train_dp},
We first compare \name with state-of-the-art approaches to compress the gradients of weights in data parallelism: 1-bit Adam and 1-bit LAMB. Both baselines achieve an average bits of 3.25,
as they require a warm-up period for the initial 15\% of training iterations where gradients remain uncompressed. Empirically, 1-bit Adam achieves a validation perplexity of 54.6, while 1-bit LAMB reaches 79.0. \namens, however, achieves 51.0 with an average of only 2.6 bits, close to that of 48.2 for uncompressed training. To further explore the limits of our method, we introduce two variants with lower bit-widths. Our method can compress the gradients to 1.4 bits with a 54.8 perplexity, which is comparable to the best baseline using an average of 3.5 bits. When we further reduce the bit-width to 0.8 bits, our method converges early with a 78.7 validation perplexity, performing on par with the 1-bit LAMB baseline but at a much lower bit-width. This demonstrates the versatility of our method, as it can trade off between compression ratios and trained model quality across a wide range of bit-widths. Moreover, it is important to note that 1-bit Adam and 1-bit LAMB replace the widely adopted Adam optimizer, resulting in significant instability during training, as evidenced by large fluctuations in training loss. In contrast, \name does not make any assumptions about the training progress, eliminates the need for a warm-up period, and maintains stability throughout training.

In addition to these baselines, we also compare our method with 2-bit and 4-bit RTN quantization with a group size of 128. Our results shows that directly quantizing gradients to 4 bits results in a perplexity of 50.2, while the 2-bit variant completely fails to converge. The compression quality ranks s follows: \name (2.6 bits) $>$ RTN (4 bits) $>$ \name (1.4 bits) $>$ \name (0.8 bits) $>$ RTN (2 bits). Since RTN quantization is also a vanilla compression algorithm, these results further demonstrate the superior compression capability of \namens.

\section{Insights for LLM Accelerator Design}
Despite \name 
achieving state-of-the-art information efficiency for compressing tensors, 
it is currently bottlenecked by the limited throughput of built-in video encoders and decoders on GPUs. 
Since people typically watch videos at resolutions lower than 4K, and at a framerate lower than 60 frames per second, the video codecs on hardware, such as the NVENC and NVDEC engines on Nvidia GPUs, lack the incentive to support higher throughput. 
These engines, when used for tensor compression in \namens, limit the training and inference throughput. 
In our measurements, NVENC achieves a throughput of around 900MB/s for compressing tensors, while NVDEC achieves a throughput of around 1100MB/s for decompressing video bitstreams to tensors, limiting the GPU's end-to-end communication bandwidth to 900MB/s.

In this section, we will delve deeper into the hardware implementation details of video codecs and propose augmentations for the design of future tensor-specialized codecs and compression-enabled training and serving systems. 
We found that video codecs are highly cost-efficient. 
Furthermore, many of their components are redundant for tensor compression, offering opportunities for further optimization.
We envision that future accelerators can trade minimal cost to implement high-throughput tensor codecs for more scalable and efficient distributed LLM training and inference.

\subsection{Build More Codecs for Future LLM Training Accelerators}
We first obtained open-sourced RTL hardware implementations of both the encoder~\cite{h264_encoder, h265_encoder} and the decoder~\cite{h264_decoder, h265_decoder} of the H.264 and the H.265 respectively.
We synthesized, placed, and routed these hardware modules using the ASAP7~\cite{asap7} 7nm technology library.
In prior sections, we have been using H.265 for all our inference and training experiments because it is more information-efficient than H.264. 
In this section, we also present the hardware evaluation of H.264, which can be considered a cheaper hardware alternative but compresses tensors with less information efficiency.

\begin{figure}
    \centering
\includegraphics[width=\columnwidth]{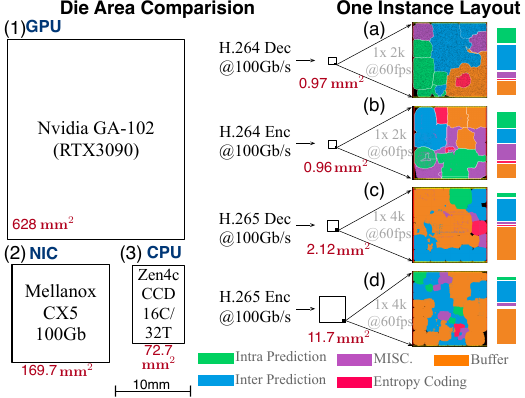}
    \caption{Comparison of the chip die area between GPU (1), CPU (3), NIC (2), and Video Codecs (a-d). Multiple instances of video encoders or decoders are combined to achieve a total throughput of 100Gb/s.}
    \label{fig:die_area_comparision}
\end{figure}

Comparisons of the die area with other devices commonly used in an LLM training data center, such as GPUs, NICs, and CPUs, are shown in Figure~\ref{fig:die_area_comparision}.
Note that a single instance of the codec supports resolutions up to $3840 \times 2160$ and throughput up to 60 frames per second. 
For fair comparisons, we normalized the throughput of the encoders and decoders to match the 100Gbps NIC bandwidth, thereby aggregating multiple instances of a video encoder or decoder in parallel to achieve the desired throughput.

Looking at the die area comparison of these devices, we observed that video codecs occupy significantly less space than other devices in data centers. 
The die area of an RTX3090 GPU is 628 mm$^2$, while a combination of an H.264 encoder and decoder, each capable of processing up to 100 Gbps, requires less than 2mm$^2$ die area, which is 314$\times$ smaller than the GPU and 85$\times$ smaller than the Mellanox CX5 100Gbps Network Card.
Building additional video encoders and decoders in GPUs to increase compression and decompression bandwidth will enable efficient compression at a lower cost.

\subsection{Video Codecs $\Rightarrow$ Tensor Codecs}

\begin{table}[]
\centering
\small
\caption{Energy for Communication vs. Compression.}
\resizebox{0.9\linewidth}{!}{
\begin{tabular}{l|l|l|l}
                        & Power & Area & Energy/Bit \\
                        & (W) & (mm$^2$) & (pJ) \\
                        \hline
                        \hline
NCCL End to End      &  - & - & 5120                \\
                        \hline
H.264 Enc (100Gbps)       & 1.1  & 0.96  & 167.8                 \\
                        \hline
H.264 Dec (100Gbps)       & 1.0 & 0.97  & 154.3                \\
                        \hline
H.265 Enc (100Gbps)       & 11.0  & 11.7  & 1707.5                \\
                        \hline
H.265 Dec (100Gbps)       & 4.3  & 2.1  & 665.4                 \\
                        \hline
T.264 Enc (100Gbps)     & 0.6   &  0.6     & 97.8                 \\
                        \hline
T.264 Dec (100Gbps)     & 0.4 &  0.4    & 63.5                \\
                        \hline
T.265 Enc (100Gbps)     & 2.3 &  2.4    & 352.9                 \\
                        \hline
T.265 Dec (100Gbps)     & 0.9 &  0.5    & 144.4                \\

                        \hline
\end{tabular}}
\label{tab:hw_impl_result}
\end{table}

In addition, as we have analyzed in Section~\ref{sec:codec_for_tensor}, not all the components in the video codecs work for tensor compression. Specifically, the inter-frame prediction does not work for tensor compression.
Although these options can be disabled by setting parameters for video codecs, if a codec specialized in tensor compression is preferred, implementing a tensor codec that removes these hardware submodules will save energy and transistors. 
The zoomed-in die layout of the encoders and decoders are shown in Figure~\ref{fig:die_area_comparision} (a-d), respectively, where the die area distribution of each component is shown. 
From this distribution, we can see that most of the die area is spent on inter-frame prediction and the frame buffer. 
If the inter-frame prediction is removed, we save the die area spent for the inter-frame prediction logic and drastically decrease the buffer size requirement as frames no longer need to be maintained for analyzing inter-frame correlations.

We modified the existing video codecs, removed the inter-frame prediction logic, and adjusted the size of the frame buffers. 
The hardware implementation results of these new tensor codecs, which we refer to as T.264 and T.265, are shown in Table~\ref{tab:hw_impl_result}.
Simplified from video codecs, these tensor codecs exhibit significantly smaller die area and lower power consumption while achieving the same throughput and compression quality as the original video codecs.

\subsection{Performance Impact of Compression}

We developed an analytical model for modeling the performance and energy consumption of a distributed training cluster to investigate the effect of enabling communication compression for future larger models. 
The model considers the LLM's configuration and GPU specifications, such as memory capacity and GPU power. 
It evaluates the GPUs' performance and power, as well as the performance and energy consumed during communication and compression. 
For each input pair of LLM's configuration and system specification, it automatically infers the best data parallelism and pipeline parallelism configuration so that the model fits into the DRAM of each GPU and the total communication size is minimized. 
To calibrate the model with realistic data, we run micro-benchmarks on servers to verify the performance and measure the power used for communication\footnote{We measured the end-to-end communication power of NCCL~\cite{nccl} as indicated in Table~\ref{tab:hw_impl_result}. NCCL tests~\cite{nccltests} was executed and power is measured using the power sensors on the Server Board Management Controllers (BMC).}.

\begin{figure}[]
    \centering
    \includegraphics[width=\columnwidth]{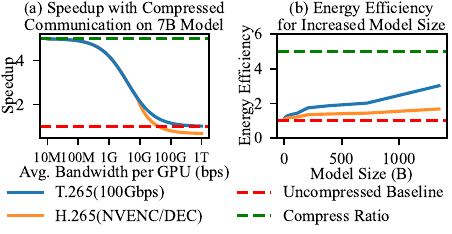}
    \caption{Impact of Communication Compression in Distributed LLM Training. The "Uncompressed Baseline" represents results without compression. The compress ratio determines \mbox{the upper bound for speedup and energy efficiency.}}
    \label{fig:scalability}
\end{figure}

We first analyze the impact of communication compression under different communication bandwidths for training a LLaMa-2-7B model. 
As shown in Figure~\ref{fig:scalability} (a), the impact of communication compression is larger when the bandwidth heavily bottlenecks the workload. 
We show two cases in this figure. 
The first case utilizes the NVENC and NVDEC engines available on Nvidia GPUs. 
Since these engines are designed for video streaming, they offer a bandwidth of only around 900MB/s and 1100MB/s.
This bandwidth is insufficient for compressing large volumes of tensors during distributed inference.
Nevertheless, the compression still substantially accelerates the training when the communication bandwidth per GPU is less than 10Gbps. 
Suppose we can afford to add a T.265 module of 3.2mm$^2$ die area onto the GPU, increasing the GPU's total die area by only 0.5\%. 
Such a minor increase in the area trade significantly accelerated distributed training with communication compression at higher bandwidth, indicating the benefits of integrating a high-bandwidth custom tensor codec into future GPU and accelerator designs.

\subsection{Compression for Scalability and Sustainability}
As models scale up, the communication bottleneck becomes more severe due to the memory constraint on a single GPU and the need to split the model into smaller parts.
The communication bottleneck not only hampers training efficiency but also translates to a higher amount of energy spent on transferring data. 
We calculate the energy consumed for encoding/decoding one byte or transmitting one bit for codecs and interfaces in Table~\ref{tab:hw_impl_result}. 
When comparing the energy used for compressing and communicating, we observed that compression requires significantly less energy. 
For example, the combined energy used for T.264 encoding and decoding is $\frac{5120}{97.8+63.5}=31.7\times$ lower than that used for end-to-end communication with NCCL. 
As shown in Section~\ref{sec:dist_training}, video codecs achieve a compression ratio of 3-20$\times$. 
For example, if a 5$\times$ compression ratio on average can be achieved, it translates to $\frac{5120}{5120 / 5 + 97.8+63.5}=4.32\times$ energy efficiency compared to transferring everything in an uncompressed format. 

The modeling of compression-enabled training at the cluster level is shown in Figure~\ref{fig:scalability} (b), where we plot the energy efficiency of using compressed communication using codecs versus increased model size. 
For distributed LLM training, communication power will account for a significant portion of the total power consumption. 
The larger the model is, the greater the percentage of power consumed by communication. 
By employing communication compression, the size of data being transmitted can be significantly reduced, resulting in power efficiency several times better than if left uncompressed.
This highlights the importance of deploying high-bandwidth customized tensor codecs on GPUs and accelerators to ensure the scalability and sustainability of data centers for training future larger and larger LLMs.

\section{Conclusion}
\name repurposed video codecs as general-purpose and versatile tensor codecs. 
Leveraging the hardware video encoding/decoding engines available on modern GPUs, \name achieves state-of-the-art information efficiency for compressing weights, activations, and gradients of LLMs. 
This greatly reduces the pressure on the memory capacity and communication bandwidth of GPUs.
To fully unlock the potential of \namens, we propose integrating specialized high throughput but cheap tensor codecs on future GPUs for more efficient distributed LLM training and inference.
\name will be open-sourced upon the acceptance of this paper.

\section*{Acknowledgment}
This work was supported in part by a National Science Foundation CAREER award NSF CCF-2045973, in part by a National Science Foundation CAREER award CNS-2238665, and in part by a National Science Foundation award NSF CNS-2112562 (NSF AI Institute - Athena). 

\raisebox{-0.3\height}{\includegraphics[height=24pt]{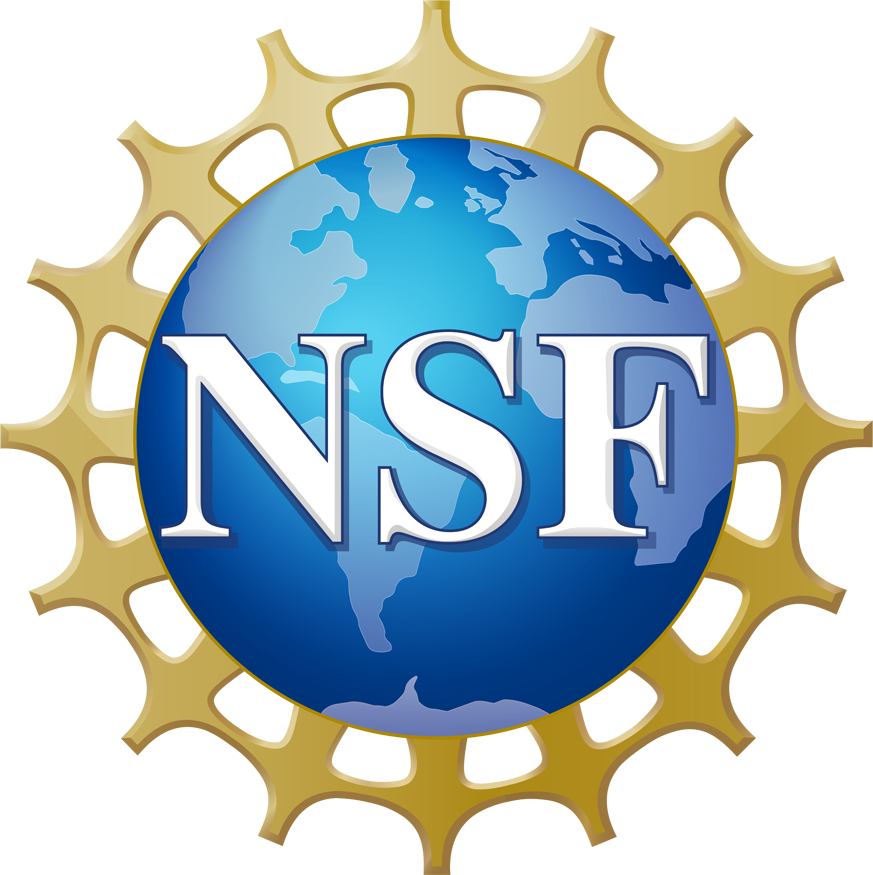}}

\bibliographystyle{IEEEtran}
\bibliography{ref}

\begin{thebibliography}{10}
\providecommand{\url}[1]{#1}
\csname url@samestyle\endcsname
\providecommand{\newblock}{\relax}
\providecommand{\bibinfo}[2]{#2}
\providecommand{\BIBentrySTDinterwordspacing}{\spaceskip=0pt\relax}
\providecommand{\BIBentryALTinterwordstretchfactor}{4}
\providecommand{\BIBentryALTinterwordspacing}{\spaceskip=\fontdimen2\font plus
\BIBentryALTinterwordstretchfactor\fontdimen3\font minus
  \fontdimen4\font\relax}
\providecommand{\BIBforeignlanguage}[2]{{%
\expandafter\ifx\csname l@#1\endcsname\relax
\typeout{** WARNING: IEEEtran.bst: No hyphenation pattern has been}%
\typeout{** loaded for the language `#1'. Using the pattern for}%
\typeout{** the default language instead.}%
\else
\language=\csname l@#1\endcsname
\fi
#2}}
\providecommand{\BIBdecl}{\relax}
\BIBdecl

\bibitem{gpt4}
OpenAI, J.~Achiam, S.~Adler, S.~Agarwal, L.~Ahmad, I.~Akkaya, F.~L. Aleman,
  D.~Almeida, J.~Altenschmidt, S.~Altman, S.~Anadkat, R.~Avila, I.~Babuschkin,
  S.~Balaji, V.~Balcom, P.~Baltescu, H.~Bao, M.~Bavarian, J.~Belgum, I.~Bello,
  J.~Berdine, G.~Bernadett-Shapiro, C.~Berner, L.~Bogdonoff, O.~Boiko, M.~Boyd,
  A.-L. Brakman, G.~Brockman, T.~Brooks, M.~Brundage, K.~Button, T.~Cai,
  R.~Campbell, A.~Cann, B.~Carey, C.~Carlson, R.~Carmichael, B.~Chan, C.~Chang,
  F.~Chantzis, D.~Chen, S.~Chen, R.~Chen, J.~Chen, M.~Chen, B.~Chess, C.~Cho,
  C.~Chu, H.~W. Chung, D.~Cummings, J.~Currier, Y.~Dai, C.~Decareaux, T.~Degry,
  N.~Deutsch, D.~Deville, A.~Dhar, D.~Dohan, S.~Dowling, S.~Dunning,
  A.~Ecoffet, A.~Eleti, T.~Eloundou, D.~Farhi, L.~Fedus, N.~Felix, S.~P.
  Fishman, J.~Forte, I.~Fulford, L.~Gao, E.~Georges, C.~Gibson, V.~Goel,
  T.~Gogineni, G.~Goh, R.~Gontijo-Lopes, J.~Gordon, M.~Grafstein, S.~Gray,
  R.~Greene, J.~Gross, S.~S. Gu, Y.~Guo, C.~Hallacy, J.~Han, J.~Harris, Y.~He,
  M.~Heaton, J.~Heidecke, C.~Hesse, A.~Hickey, W.~Hickey, P.~Hoeschele,
  B.~Houghton, K.~Hsu, S.~Hu, X.~Hu, J.~Huizinga, S.~Jain, S.~Jain, J.~Jang,
  A.~Jiang, R.~Jiang, H.~Jin, D.~Jin, S.~Jomoto, B.~Jonn, H.~Jun, T.~Kaftan,
  Łukasz Kaiser, A.~Kamali, I.~Kanitscheider, N.~S. Keskar, T.~Khan,
  L.~Kilpatrick, J.~W. Kim, C.~Kim, Y.~Kim, J.~H. Kirchner, J.~Kiros,
  M.~Knight, D.~Kokotajlo, Łukasz Kondraciuk, A.~Kondrich, A.~Konstantinidis,
  K.~Kosic, G.~Krueger, V.~Kuo, M.~Lampe, I.~Lan, T.~Lee, J.~Leike, J.~Leung,
  D.~Levy, C.~M. Li, R.~Lim, M.~Lin, S.~Lin, M.~Litwin, T.~Lopez, R.~Lowe,
  P.~Lue, A.~Makanju, K.~Malfacini, S.~Manning, T.~Markov, Y.~Markovski,
  B.~Martin, K.~Mayer, A.~Mayne, B.~McGrew, S.~M. McKinney, C.~McLeavey,
  P.~McMillan, J.~McNeil, D.~Medina, A.~Mehta, J.~Menick, L.~Metz,
  A.~Mishchenko, P.~Mishkin, V.~Monaco, E.~Morikawa, D.~Mossing, T.~Mu,
  M.~Murati, O.~Murk, D.~Mély, A.~Nair, R.~Nakano, R.~Nayak, A.~Neelakantan,
  R.~Ngo, H.~Noh, L.~Ouyang, C.~O'Keefe, J.~Pachocki, A.~Paino, J.~Palermo,
  A.~Pantuliano, G.~Parascandolo, J.~Parish, E.~Parparita, A.~Passos,
  M.~Pavlov, A.~Peng, A.~Perelman, F.~de~Avila Belbute~Peres, M.~Petrov, H.~P.
  de~Oliveira~Pinto, Michael, Pokorny, M.~Pokrass, V.~H. Pong, T.~Powell,
  A.~Power, B.~Power, E.~Proehl, R.~Puri, A.~Radford, J.~Rae, A.~Ramesh,
  C.~Raymond, F.~Real, K.~Rimbach, C.~Ross, B.~Rotsted, H.~Roussez, N.~Ryder,
  M.~Saltarelli, T.~Sanders, S.~Santurkar, G.~Sastry, H.~Schmidt, D.~Schnurr,
  J.~Schulman, D.~Selsam, K.~Sheppard, T.~Sherbakov, J.~Shieh, S.~Shoker,
  P.~Shyam, S.~Sidor, E.~Sigler, M.~Simens, J.~Sitkin, K.~Slama, I.~Sohl,
  B.~Sokolowsky, Y.~Song, N.~Staudacher, F.~P. Such, N.~Summers, I.~Sutskever,
  J.~Tang, N.~Tezak, M.~B. Thompson, P.~Tillet, A.~Tootoonchian, E.~Tseng,
  P.~Tuggle, N.~Turley, J.~Tworek, J.~F.~C. Uribe, A.~Vallone, A.~Vijayvergiya,
  C.~Voss, C.~Wainwright, J.~J. Wang, A.~Wang, B.~Wang, J.~Ward, J.~Wei,
  C.~Weinmann, A.~Welihinda, P.~Welinder, J.~Weng, L.~Weng, M.~Wiethoff,
  D.~Willner, C.~Winter, S.~Wolrich, H.~Wong, L.~Workman, S.~Wu, J.~Wu, M.~Wu,
  K.~Xiao, T.~Xu, S.~Yoo, K.~Yu, Q.~Yuan, W.~Zaremba, R.~Zellers, C.~Zhang,
  M.~Zhang, S.~Zhao, T.~Zheng, J.~Zhuang, W.~Zhuk, and B.~Zoph, ``Gpt-4
  technical report,'' 2024.

\bibitem{langchian_custom_service}
K.~Pandya and M.~Holia, ``Automating customer service using langchain: Building
  custom open-source gpt chatbot for organizations,'' \emph{arXiv preprint
  arXiv:2310.05421}, 2023.

\bibitem{emergent}
J.~Wei, Y.~Tay, R.~Bommasani, C.~Raffel, B.~Zoph, S.~Borgeaud, D.~Yogatama,
  M.~Bosma, D.~Zhou, D.~Metzler \emph{et~al.}, ``Emergent abilities of large
  language models,'' \emph{arXiv preprint arXiv:2206.07682}, 2022.

\bibitem{starcoder}
A.~Lozhkov, R.~Li, L.~B. Allal, F.~Cassano, J.~Lamy-Poirier, N.~Tazi, A.~Tang,
  D.~Pykhtar, J.~Liu, Y.~Wei, T.~Liu, M.~Tian, D.~Kocetkov, A.~Zucker,
  Y.~Belkada, Z.~Wang, Q.~Liu, D.~Abulkhanov, I.~Paul, Z.~Li, W.-D. Li,
  M.~Risdal, J.~Li, J.~Zhu, T.~Y. Zhuo, E.~Zheltonozhskii, N.~O.~O. Dade,
  W.~Yu, L.~Krauß, N.~Jain, Y.~Su, X.~He, M.~Dey, E.~Abati, Y.~Chai,
  N.~Muennighoff, X.~Tang, M.~Oblokulov, C.~Akiki, M.~Marone, C.~Mou,
  M.~Mishra, A.~Gu, B.~Hui, T.~Dao, A.~Zebaze, O.~Dehaene, N.~Patry, C.~Xu,
  J.~McAuley, H.~Hu, T.~Scholak, S.~Paquet, J.~Robinson, C.~J. Anderson,
  N.~Chapados, M.~Patwary, N.~Tajbakhsh, Y.~Jernite, C.~M. Ferrandis, L.~Zhang,
  S.~Hughes, T.~Wolf, A.~Guha, L.~von Werra, and H.~de~Vries, ``Starcoder 2 and
  the stack v2: The next generation,'' 2024.

\bibitem{codellama}
B.~Rozière, J.~Gehring, F.~Gloeckle, S.~Sootla, I.~Gat, X.~E. Tan, Y.~Adi,
  J.~Liu, R.~Sauvestre, T.~Remez, J.~Rapin, A.~Kozhevnikov, I.~Evtimov,
  J.~Bitton, M.~Bhatt, C.~C. Ferrer, A.~Grattafiori, W.~Xiong, A.~Défossez,
  J.~Copet, F.~Azhar, H.~Touvron, L.~Martin, N.~Usunier, T.~Scialom, and
  G.~Synnaeve, ``Code llama: Open foundation models for code,'' 2024.

\bibitem{mathqa}
M.~Schubotz, P.~Scharpf, K.~Dudhat, Y.~Nagar, F.~Hamborg, and B.~Gipp,
  ``Introducing mathqa: a math-aware question answering system,''
  \emph{Information Discovery and Delivery}, vol.~46, no.~4, pp. 214--224,
  2018.

\bibitem{gpt_theory_prove}
S.~Polu and I.~Sutskever, ``Generative language modeling for automated theorem
  proving,'' \emph{arXiv preprint arXiv:2009.03393}, 2020.

\bibitem{nemotron}
N.~Inc., ``Nemotron-4 340b technical report,'' 2024.

\bibitem{llama3}
\BIBentryALTinterwordspacing
AI@Meta, ``Llama 3 model card,'' 2024. [Online]. Available:
  \url{https://github.com/meta-llama/llama3/blob/main/MODEL_CARD.md}
\BIBentrySTDinterwordspacing

\bibitem{harlap2018pipedream}
A.~Harlap, D.~Narayanan, A.~Phanishayee, V.~Seshadri, N.~Devanur, G.~Ganger,
  and P.~Gibbons, ``Pipedream: Fast and efficient pipeline parallel dnn
  training,'' \emph{arXiv preprint arXiv:1806.03377}, 2018.

\bibitem{huang2019gpipe}
Y.~Huang, Y.~Cheng, A.~Bapna, O.~Firat, D.~Chen, M.~Chen, H.~Lee, J.~Ngiam,
  Q.~V. Le, Y.~Wu \emph{et~al.}, ``Gpipe: Efficient training of giant neural
  networks using pipeline parallelism,'' \emph{Advances in neural information
  processing systems}, vol.~32, 2019.

\bibitem{dean2012large}
J.~Dean, G.~Corrado, R.~Monga, K.~Chen, M.~Devin, M.~Mao, M.~Ranzato,
  A.~Senior, P.~Tucker, K.~Yang \emph{et~al.}, ``Large scale distributed deep
  networks,'' \emph{Advances in neural information processing systems},
  vol.~25, 2012.

\bibitem{chen2016revisiting}
J.~Chen, X.~Pan, R.~Monga, S.~Bengio, and R.~Jozefowicz, ``Revisiting
  distributed synchronous sgd,'' \emph{arXiv preprint arXiv:1604.00981}, 2016.

\bibitem{sun2024trustllm}
L.~Sun, Y.~Huang, H.~Wang, S.~Wu, Q.~Zhang, C.~Gao, Y.~Huang, W.~Lyu, Y.~Zhang,
  X.~Li \emph{et~al.}, ``Trustllm: Trustworthiness in large language models,''
  \emph{arXiv preprint arXiv:2401.05561}, 2024.

\bibitem{yao2024survey}
Y.~Yao, J.~Duan, K.~Xu, Y.~Cai, Z.~Sun, and Y.~Zhang, ``A survey on large
  language model (llm) security and privacy: The good, the bad, and the ugly,''
  \emph{High-Confidence Computing}, p. 100211, 2024.

\bibitem{liang2024communication}
F.~Liang, Z.~Zhang, H.~Lu, V.~Leung, Y.~Guo, and X.~Hu,
  ``Communication-efficient large-scale distributed deep learning: A
  comprehensive survey,'' \emph{arXiv preprint arXiv:2404.06114}, 2024.

\bibitem{thorpe2023bamboo}
J.~Thorpe, P.~Zhao, J.~Eyolfson, Y.~Qiao, Z.~Jia, M.~Zhang, R.~Netravali, and
  G.~H. Xu, ``Bamboo: Making preemptible instances resilient for affordable
  training of large $\{$DNNs$\}$,'' in \emph{20th USENIX Symposium on Networked
  Systems Design and Implementation (NSDI 23)}, 2023, pp. 497--513.

\bibitem{patel_nishball_2024}
\BIBentryALTinterwordspacing
D.~Patel and D.~Nishball, ``100,000 h100 clusters: Power, network topology,
  ethernet vs infiniband, reliability, failures, checkpointing,''
  \emph{SemiAnalysis}, 2024, accessed: 2024-06-22. [Online]. Available:
  \url{https://www.semianalysis.com/p/100000-h100-clusters-power-network}
\BIBentrySTDinterwordspacing

\bibitem{tpuv4}
\BIBentryALTinterwordspacing
N.~Jouppi, G.~Kurian, S.~Li, P.~Ma, R.~Nagarajan, L.~Nai, N.~Patil,
  S.~Subramanian, A.~Swing, B.~Towles, C.~Young, X.~Zhou, Z.~Zhou, and D.~A.
  Patterson, ``Tpu v4: An optically reconfigurable supercomputer for machine
  learning with hardware support for embeddings,'' in \emph{Proceedings of the
  50th Annual International Symposium on Computer Architecture}, ser. ISCA
  '23.\hskip 1em plus 0.5em minus 0.4em\relax New York, NY, USA: Association
  for Computing Machinery, 2023. [Online]. Available:
  \url{https://doi.org/10.1145/3579371.3589350}
\BIBentrySTDinterwordspacing

\bibitem{gptq}
E.~Frantar, S.~Ashkboos, T.~Hoefler, and D.~Alistarh, ``Gptq: Accurate
  post-training quantization for generative pre-trained transformers,'' 2023.

\bibitem{awq}
J.~Lin, J.~Tang, H.~Tang, S.~Yang, W.-M. Chen, W.-C. Wang, G.~Xiao, X.~Dang,
  C.~Gan, and S.~Han, ``Awq: Activation-aware weight quantization for llm
  compression and acceleration,'' 2024.

\bibitem{adam1bit}
H.~Tang, S.~Gan, A.~A. Awan, S.~Rajbhandari, C.~Li, X.~Lian, J.~Liu, C.~Zhang,
  and Y.~He, ``1-bit adam: Communication efficient large-scale training with
  adam's convergence speed,'' 2021.

\bibitem{lamb1bit}
C.~Li, A.~A. Awan, H.~Tang, S.~Rajbhandari, and Y.~He, ``1-bit lamb:
  Communication efficient large-scale large-batch training with lamb's
  convergence speed,'' 2021.

\bibitem{smoothquant}
G.~Xiao, J.~Lin, M.~Seznec, H.~Wu, J.~Demouth, and S.~Han, ``Smoothquant:
  Accurate and efficient post-training quantization for large language
  models,'' 2024.

\bibitem{videosdk}
``{NVIDIA Video Codec SDK},''
  \url{https://developer.nvidia.com/nvidia-video-codec-sdk/download}, 2024.

\bibitem{han2015deep}
S.~Han, H.~Mao, and W.~J. Dally, ``Deep compression: Compressing deep neural
  networks with pruning, trained quantization and huffman coding,'' \emph{arXiv
  preprint arXiv:1510.00149}, 2015.

\bibitem{Jacob_2018_CVPR}
B.~Jacob, S.~Kligys, B.~Chen, M.~Zhu, M.~Tang, A.~Howard, H.~Adam, and
  D.~Kalenichenko, ``Quantization and training of neural networks for efficient
  integer-arithmetic-only inference,'' in \emph{Proceedings of the IEEE
  Conference on Computer Vision and Pattern Recognition (CVPR)}, June 2018.

\bibitem{banner2018scalable}
R.~Banner, I.~Hubara, E.~Hoffer, and D.~Soudry, ``Scalable methods for 8-bit
  training of neural networks,'' \emph{Advances in neural information
  processing systems}, vol.~31, 2018.

\bibitem{nagel2019data}
M.~Nagel, M.~v. Baalen, T.~Blankevoort, and M.~Welling, ``Data-free
  quantization through weight equalization and bias correction,'' in
  \emph{Proceedings of the IEEE/CVF International Conference on Computer
  Vision}, 2019, pp. 1325--1334.

\bibitem{kim2023squeezellm}
S.~Kim, C.~Hooper, A.~Gholami, Z.~Dong, X.~Li, S.~Shen, M.~W. Mahoney, and
  K.~Keutzer, ``Squeezellm: Dense-and-sparse quantization,'' \emph{arXiv
  preprint arXiv:2306.07629}, 2023.

\bibitem{tseng2024quip}
A.~Tseng, J.~Chee, Q.~Sun, V.~Kuleshov, and C.~De~Sa, ``Quip\#: Even better llm
  quantization with hadamard incoherence and lattice codebooks,'' \emph{arXiv
  preprint arXiv:2402.04396}, 2024.

\bibitem{qlora}
T.~Dettmers, A.~Pagnoni, A.~Holtzman, and L.~Zettlemoyer, ``Qlora: Efficient
  finetuning of quantized llms,'' 2023.

\bibitem{zhang2024h2o}
Z.~Zhang, Y.~Sheng, T.~Zhou, T.~Chen, L.~Zheng, R.~Cai, Z.~Song, Y.~Tian,
  C.~R{\'e}, C.~Barrett \emph{et~al.}, ``H2o: Heavy-hitter oracle for efficient
  generative inference of large language models,'' \emph{Advances in Neural
  Information Processing Systems}, vol.~36, 2024.

\bibitem{song2023optimus}
J.~Song, J.~Yim, J.~Jung, H.~Jang, H.-J. Kim, Y.~Kim, and J.~Lee, ``Optimus-cc:
  Efficient large nlp model training with 3d parallelism aware communication
  compression,'' in \emph{Proceedings of the 28th ACM International Conference
  on Architectural Support for Programming Languages and Operating Systems,
  Volume 2}, 2023, pp. 560--573.

\bibitem{sun2023simple}
M.~Sun, Z.~Liu, A.~Bair, and J.~Z. Kolter, ``A simple and effective pruning
  approach for large language models,'' \emph{arXiv preprint arXiv:2306.11695},
  2023.

\bibitem{liu2023llm}
Z.~Liu, B.~Oguz, C.~Zhao, E.~Chang, P.~Stock, Y.~Mehdad, Y.~Shi,
  R.~Krishnamoorthi, and V.~Chandra, ``Llm-qat: Data-free quantization aware
  training for large language models,'' \emph{arXiv preprint arXiv:2305.17888},
  2023.

\bibitem{xu2024onebit}
Y.~Xu, X.~Han, Z.~Yang, S.~Wang, Q.~Zhu, Z.~Liu, W.~Liu, and W.~Che, ``Onebit:
  Towards extremely low-bit large language models,'' \emph{arXiv preprint
  arXiv:2402.11295}, 2024.

\bibitem{quip}
J.~Chee, Y.~Cai, V.~Kuleshov, and C.~M. De~Sa, ``Quip: 2-bit quantization of
  large language models with guarantees,'' \emph{Advances in Neural Information
  Processing Systems}, vol.~36, 2024.

\bibitem{qserve}
Y.~Lin, H.~Tang, S.~Yang, Z.~Zhang, G.~Xiao, C.~Gan, and S.~Han, ``Qserve:
  W4a8kv4 quantization and system co-design for efficient llm serving,'' 2024.

\bibitem{quarot}
S.~Ashkboos, A.~Mohtashami, M.~L. Croci, B.~Li, M.~Jaggi, D.~Alistarh,
  T.~Hoefler, and J.~Hensman, ``Quarot: Outlier-free 4-bit inference in rotated
  llms,'' 2024.

\bibitem{spinquant}
Z.~Liu, C.~Zhao, I.~Fedorov, B.~Soran, D.~Choudhary, R.~Krishnamoorthi,
  V.~Chandra, Y.~Tian, and T.~Blankevoort, ``Spinquant--llm quantization with
  learned rotations,'' \emph{arXiv preprint arXiv:2405.16406}, 2024.

\bibitem{h264_standard}
\BIBentryALTinterwordspacing
{International Telecommunication Union}, ``{ITU-T Recommendation H.264:
  Advanced Video Coding for Generic Audiovisual Services},'' {International
  Telecommunication Union}, Tech. Rep., 2023. [Online]. Available:
  \url{https://www.itu.int/rec/T-REC-H.264}
\BIBentrySTDinterwordspacing

\bibitem{h265_standard}
\BIBentryALTinterwordspacing
------, ``{ITU-T Recommendation H.265: High Efficiency Video Coding},''
  {International Telecommunication Union}, Tech. Rep., 2023. [Online].
  Available: \url{https://www.itu.int/rec/T-REC-H.265}
\BIBentrySTDinterwordspacing

\bibitem{hevc_overview}
G.~J. Sullivan, J.-R. Ohm, W.-J. Han, and T.~Wiegand, ``Overview of the high
  efficiency video coding (hevc) standard,'' \emph{IEEE Transactions on
  Circuits and Systems for Video Technology}, vol.~22, no.~12, pp. 1649--1668,
  2012.

\bibitem{cabac}
D.~Marpe, H.~Schwarz, and T.~Wiegand, ``Context-based adaptive binary
  arithmetic coding in the h.264/avc video compression standard,'' \emph{IEEE
  Transactions on Circuits and Systems for Video Technology}, vol.~13, no.~7,
  pp. 620--636, 2003.

\bibitem{llama2}
H.~Touvron, L.~Martin, K.~Stone, P.~Albert, A.~Almahairi, Y.~Babaei,
  N.~Bashlykov, S.~Batra, P.~Bhargava, S.~Bhosale \emph{et~al.}, ``Llama 2:
  Open foundation and fine-tuned chat models,'' \emph{arXiv preprint
  arXiv:2307.09288}, 2023.

\bibitem{fp4}
\BIBentryALTinterwordspacing
S.-y. Liu, Z.~Liu, X.~Huang, P.~Dong, and K.-T. Cheng, ``{LLM}-{FP}4: 4-bit
  floating-point quantized transformers,'' in \emph{Proceedings of the 2023
  Conference on Empirical Methods in Natural Language Processing}, H.~Bouamor,
  J.~Pino, and K.~Bali, Eds.\hskip 1em plus 0.5em minus 0.4em\relax Singapore:
  Association for Computational Linguistics, Dec. 2023, pp. 592--605. [Online].
  Available: \url{https://aclanthology.org/2023.emnlp-main.39}
\BIBentrySTDinterwordspacing

\bibitem{dct}
S.~A. Khayam, ``The discrete cosine transform (dct): theory and application,''
  \emph{Michigan State University}, vol. 114, no.~1, p.~31, 2003.

\bibitem{slicegpt}
S.~Ashkboos, M.~L. Croci, M.~G. do~Nascimento, T.~Hoefler, and J.~Hensman,
  ``Slicegpt: Compress large language models by deleting rows and columns,''
  2024.

\bibitem{llmint8}
T.~Dettmers, M.~Lewis, Y.~Belkada, and L.~Zettlemoyer, ``Llm.int8(): 8-bit
  matrix multiplication for transformers at scale,'' 2022.

\bibitem{hm}
\BIBentryALTinterwordspacing
{International Telecommunication Union}, ``Hevc test model (hm),'' Tech. Rep.,
  2023. [Online]. Available: \url{https://hevc.hhi.fraunhofer.de/}
\BIBentrySTDinterwordspacing

\bibitem{pytorch}
J.~Ansel, E.~Yang, H.~He, N.~Gimelshein, A.~Jain, M.~Voznesensky, B.~Bao,
  P.~Bell, D.~Berard, E.~Burovski \emph{et~al.}, ``Pytorch 2: Faster machine
  learning through dynamic python bytecode transformation and graph
  compilation,'' 2024.

\bibitem{outliers}
D.~Paglieri, S.~Dash, T.~Rockt{\"a}schel, and J.~Parker-Holder, ``Outliers and
  calibration sets have diminishing effect on quantization of modern llms,''
  \emph{arXiv preprint arXiv:2405.20835}, 2024.

\bibitem{eval-harness}
\BIBentryALTinterwordspacing
L.~Gao, J.~Tow, B.~Abbasi, S.~Biderman, S.~Black, A.~DiPofi, C.~Foster,
  L.~Golding, J.~Hsu, A.~Le~Noac'h, H.~Li, K.~McDonell, N.~Muennighoff,
  C.~Ociepa, J.~Phang, L.~Reynolds, H.~Schoelkopf, A.~Skowron, L.~Sutawika,
  E.~Tang, A.~Thite, B.~Wang, K.~Wang, and A.~Zou, ``A framework for few-shot
  language model evaluation,'' 12 2023. [Online]. Available:
  \url{https://zenodo.org/records/10256836}
\BIBentrySTDinterwordspacing

\bibitem{piqa}
Y.~Bisk, R.~Zellers, J.~Gao, Y.~Choi \emph{et~al.}, ``Piqa: Reasoning about
  physical commonsense in natural language,'' in \emph{Proceedings of the AAAI
  conference on artificial intelligence}, vol.~34, no.~05, 2020, pp.
  7432--7439.

\bibitem{copa}
M.~Roemmele, C.~A. Bejan, and A.~S. Gordon, ``Choice of plausible alternatives:
  An evaluation of commonsense causal reasoning,'' in \emph{2011 AAAI Spring
  Symposium Series}, 2011.

\bibitem{clark2018think}
P.~Clark, I.~Cowhey, O.~Etzioni, T.~Khot, A.~Sabharwal, C.~Schoenick, and
  O.~Tafjord, ``Think you have solved question answering? try arc, the ai2
  reasoning challenge,'' \emph{arXiv preprint arXiv:1803.05457}, 2018.

\bibitem{sakaguchi2021winogrande}
K.~Sakaguchi, R.~L. Bras, C.~Bhagavatula, and Y.~Choi, ``Winogrande: An
  adversarial winograd schema challenge at scale,'' \emph{Communications of the
  ACM}, vol.~64, no.~9, pp. 99--106, 2021.

\bibitem{zellers2019hellaswag}
R.~Zellers, A.~Holtzman, Y.~Bisk, A.~Farhadi, and Y.~Choi, ``Hellaswag: Can a
  machine really finish your sentence?'' \emph{arXiv preprint
  arXiv:1905.07830}, 2019.

\bibitem{glue}
A.~Wang, A.~Singh, J.~Michael, F.~Hill, O.~Levy, and S.~R. Bowman, ``Glue: A
  multi-task benchmark and analysis platform for natural language
  understanding,'' \emph{arXiv preprint arXiv:1804.07461}, 2018.

\bibitem{mihaylov2018can}
T.~Mihaylov, P.~Clark, T.~Khot, and A.~Sabharwal, ``Can a suit of armor conduct
  electricity? a new dataset for open book question answering,'' \emph{arXiv
  preprint arXiv:1809.02789}, 2018.

\bibitem{merity2016pointer}
S.~Merity, C.~Xiong, J.~Bradbury, and R.~Socher, ``Pointer sentinel mixture
  models,'' 2016.

\bibitem{sun2024triforce}
H.~Sun, Z.~Chen, X.~Yang, Y.~Tian, and B.~Chen, ``Triforce: Lossless
  acceleration of long sequence generation with hierarchical speculative
  decoding,'' \emph{arXiv preprint arXiv:2404.11912}, 2024.

\bibitem{dong2024get}
H.~Dong, X.~Yang, Z.~Zhang, Z.~Wang, Y.~Chi, and B.~Chen, ``Get more with less:
  Synthesizing recurrence with kv cache compression for efficient llm
  inference,'' \emph{arXiv preprint arXiv:2402.09398}, 2024.

\bibitem{wang2023topoopt}
W.~Wang, M.~Khazraee, Z.~Zhong, M.~Ghobadi, Z.~Jia, D.~Mudigere, Y.~Zhang, and
  A.~Kewitsch, ``$\{$TopoOpt$\}$: Co-optimizing network topology and
  parallelization strategy for distributed training jobs,'' in \emph{20th
  USENIX Symposium on Networked Systems Design and Implementation (NSDI 23)},
  2023, pp. 739--767.

\bibitem{wang2022overlap}
S.~Wang, J.~Wei, A.~Sabne, A.~Davis, B.~Ilbeyi, B.~Hechtman, D.~Chen, K.~S.
  Murthy, M.~Maggioni, Q.~Zhang \emph{et~al.}, ``Overlap communication with
  dependent computation via decomposition in large deep learning models,'' in
  \emph{Proceedings of the 28th ACM International Conference on Architectural
  Support for Programming Languages and Operating Systems, Volume 1}, 2022, pp.
  93--106.

\bibitem{rasley2020deepspeed}
J.~Rasley, S.~Rajbhandari, O.~Ruwase, and Y.~He, ``Deepspeed: System
  optimizations enable training deep learning models with over 100 billion
  parameters,'' in \emph{Proceedings of the 26th ACM SIGKDD International
  Conference on Knowledge Discovery \& Data Mining}, 2020, pp. 3505--3506.

\bibitem{biderman2023pythia}
S.~Biderman, H.~Schoelkopf, Q.~G. Anthony, H.~Bradley, K.~O’Brien,
  E.~Hallahan, M.~A. Khan, S.~Purohit, U.~S. Prashanth, E.~Raff \emph{et~al.},
  ``Pythia: A suite for analyzing large language models across training and
  scaling,'' in \emph{International Conference on Machine Learning}.\hskip 1em
  plus 0.5em minus 0.4em\relax PMLR, 2023, pp. 2397--2430.

\bibitem{gao2020pile}
L.~Gao, S.~Biderman, S.~Black, L.~Golding, T.~Hoppe, C.~Foster, J.~Phang,
  H.~He, A.~Thite, N.~Nabeshima \emph{et~al.}, ``The pile: An 800gb dataset of
  diverse text for language modeling,'' \emph{arXiv preprint arXiv:2101.00027},
  2020.

\bibitem{lin2017deep}
Y.~Lin, S.~Han, H.~Mao, Y.~Wang, and W.~J. Dally, ``Deep gradient compression:
  Reducing the communication bandwidth for distributed training,'' \emph{arXiv
  preprint arXiv:1712.01887}, 2017.

\bibitem{zhang2019gradient}
J.~Zhang, T.~He, S.~Sra, and A.~Jadbabaie, ``Why gradient clipping accelerates
  training: A theoretical justification for adaptivity,'' \emph{arXiv preprint
  arXiv:1905.11881}, 2019.

\bibitem{h264_encoder}
\BIBentryALTinterwordspacing
Y.~Fan, ``H.264 video encoder ip core,'' 2023. [Online]. Available:
  \url{https://github.com/openasic-org/xk264}
\BIBentrySTDinterwordspacing

\bibitem{h265_encoder}
\BIBentryALTinterwordspacing
------, ``H.265 video encoder ip core,'' 2023. [Online]. Available:
  \url{https://github.com/openasic-org/xk265}
\BIBentrySTDinterwordspacing

\bibitem{h264_decoder}
\BIBentryALTinterwordspacing
OsenLogic, ``Osen loigc osd10 h.264/avc baseline video decoder,'' 2024.
  [Online]. Available:
  \url{https://github.com/ICscholar/H264\_decoder-verilog-Cpp}
\BIBentrySTDinterwordspacing

\bibitem{h265_decoder}
\BIBentryALTinterwordspacing
T.~Shi, ``H265 decoder write in verilog, verified on xilinx zynq7035,'' 2024.
  [Online]. Available: \url{https://github.com/tishi43/h265\_decoder}
\BIBentrySTDinterwordspacing

\bibitem{asap7}
L.~T. Clark, V.~Vashishtha, D.~M. Harris, S.~Dietrich, and Z.~Wang, ``Design
  flows and collateral for the asap7 7nm finfet predictive process design
  kit,'' in \emph{2017 IEEE International Conference on Microelectronic Systems
  Education (MSE)}, 2017, pp. 1--4.

\bibitem{nccl}
``{The NVIDIA Collective Communication Library (NCCL)},''
  \url{https://developer.nvidia.com/nccl}, 2024.

\bibitem{nccltests}
``{NCCL Tests},'' \url{https://github.com/NVIDIA/nccl-tests}, 2024.

\end{thebibliography}

\end{document}